\documentclass[journal]{IEEEtran}

\ifCLASSINFOpdf
  \usepackage[pdftex]{graphicx}
\else
\fi

\ifCLASSOPTIONcompsoc
  \usepackage[caption=false,font=normalsize,labelfont=sf,textfont=sf]{subfig}
\else
  \usepackage[caption=false,font=footnotesize]{subfig}
\fi

\newcommand\nparticipants{28}
\newcommand\nvideos{43}

\newcommand\videostartprocessseconds{20}
\newcommand\IOU{0.1}
\newcommand\segmentmatchrate{0.5}

\begin{document}

\title{Cheating Detection Pipeline \\ for Online Interviews and Exams}

\author{\IEEEauthorblockN{Azmi Can Özgen, Mahiye Uluyağmur Öztürk, Umut Bayraktar} \\
    \IEEEauthorblockA{Huawei Turkey R\&D Center, Istanbul, Turkey}}



\maketitle

\begin{abstract}
    Remote examination and job interviews have gained popularity and become indispensable because of both pandemics and the advantage of remote working circumstances. Most companies and academic institutions utilize these systems for their recruitment processes and also for online exams. However, one of the critical problems of the remote examination systems is conducting the exams in a reliable environment. In this work, we present a cheating analysis pipeline for online interviews and exams. The system only requires a video of the candidate, which is recorded during the exam. Then cheating detection pipeline is employed to detect another person, electronic device usage, and candidate absence status. The pipeline consists of face detection, face recognition, object detection, and face tracking algorithms. To evaluate the performance of the pipeline we collected a private video dataset. The video dataset includes both cheating activities and clean videos. Ultimately, our pipeline presents an efficient and fast guideline to detect and analyze cheating activities in an online interview and exam video.
\end{abstract}


%
\IEEEpeerreviewmaketitle

\section{Introduction}

    \IEEEPARstart{I}{n} today's world, job interviews are more and more preferred to be online. One reason for this type of recruitment is pandemic, but another important reason both for employees and employers is to be able to use time and energy efficiently. Instead of taking interview exams in the physical offices, currently, the candidates can take interviews anytime and anywhere in the world by only becoming online. In this way, online interviews provide faster and easier hiring.
    
    Exams are common tools that enable people to measure their knowledge of certain subjects. Therefore, proctoring online exams in a reliable environment is very crucial for the exams to be valid. According to \cite{echeating}, about 74\% of the students think that they can easily cheat in online exams. When exams are done under conventional circumstances, candidates are to be monitored by a human proctor which is widely used for providing reliable exams. However, proctoring is a humdrum and time-consuming job for a person. It only requires watching over an individual or a group of people during an exam and prevent pre-determined cheating activities. Even if the pandemic is over, automating the proctoring process at universities or colleges can enable qualified personnel to work more efficiently. Essentially, automated proctoring systems are valuable contributions to online interview and exam systems. 
    
    During the pandemics, many exams, even chess games are played online. According to Bilen and Matros \cite{chess}, to prevent cheating activities during a chess game, and to assure fair play the best solution is to use a camera that records each players' computer screen and room.
    
    Online interview systems that record candidate's videos are generally using computer vision and machine learning algorithms to verify the validity of the interview \cite{auto_proctor,intelligent,proctor_m}. Face verification is an essential part of these systems. Besides, we believe that object detection is a necessity to track other abnormalities that may occur during the interview. It helps to detect the appearance of another's body or forbidden usage of an electronic device.
    
    One might argue that these cheating events do not have to happen in front of the cameras and may never be detectable from the video record. Although that is a perfectly sensible objection, our system only aims to process video records. And we believe that it would be beneficial to use it as a part of a broader interview system so that it has a deterrent effect on candidates from cheating.
    
\begin{figure}[h]
        \centering
        \includegraphics[width=0.99\linewidth, height=4cm]{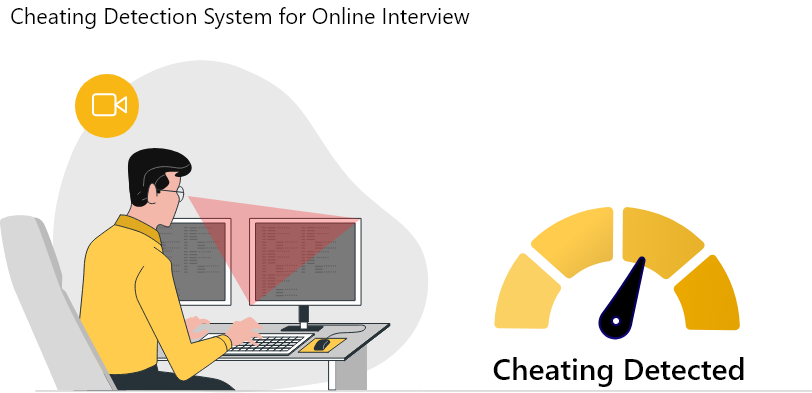}
        \caption{Cheating detection system}
        \label{fig:system}
    \end{figure}

    We built an online interview anti-cheating system supported with state-of-the-art deep learning detection models. Such a system can be used in a face-to-face interview where the interviewer and candidate can see each other, or in an online exam where the candidate answers previously prepared questions. Either way, the candidate's frontal view should be recorded as shown in Figure \ref{fig:system} for further analysis.
    
\begin{figure}[h]
        \centering
        \includegraphics[width=0.80\linewidth, height=5cm]{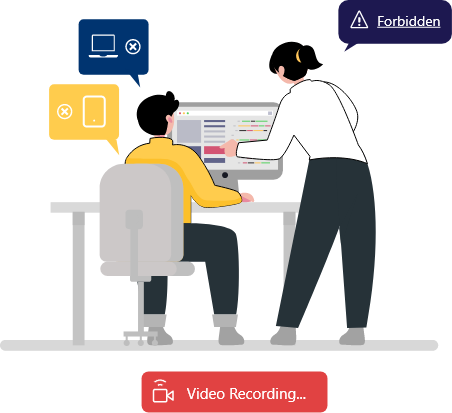}
        \caption{An exam environment with forbidden interactions.}
        \label{fig:exam_env}
    \end{figure}

    Our aim is to detect the following events that may occur during the exam: (1) presence of another person, (2) usage of mobile phone or any other electronic device, and (3) absence status. Figure \ref{fig:exam_env} shows a sample environment with the candidate and possible forbidden interactions and events that they must avoid for a legitimate interview or exam.
    
    Our contributions can be summarized as follows: 
    \begin{itemize}
    \item We have designed an efficient cheating detection pipeline.
    \item We have used a mixture of image/video processing techniques particularly developed to detect cheating activities.
    \item We have proposed cheating evaluation components such as another person's appearance, electronic device usage, and candidate absence status to decide a cheating activity.
    \item We have employed state-of-the-art deep learning models for face and object detection as cheating detection components.
    \end{itemize}

    \subsection{Related works}
    
        Researches like \cite{examsoft,comparisons} present a detailed comparison of online proctoring systems and their features.
        
        In \cite{thwart_cheating,secure_online_exam,opgen} authors propose online exam environments that can prevent cheating actions with preemptive procedures, so that without any proctor they can minimize the number of cheating attempts. In another variation of this approach, preemptive systems are combined with human proctors as in \cite{research_app,enhanced_security}
        
        There are automated interview systems that do not have AI-based anti-cheating services like \cite{codility}, \cite{HackerRank}. These systems were designed for an online coding platform to evaluate the candidates. Although \cite{codility} and \cite{HackerRank} have plagiarism services, there is no detection for the presence of multiple persons or electronic device usage.
        
        Some approaches like \cite{moop} combine autonomous detection systems with human proctors with well-defined pipelines. \cite{proctor_m} propose a pipeline that has dataset collecting and training steps for further verification of the user.
        
        There are multi-modal approaches that assemble several cheating analysis paradigms in visual and audial features. \cite{hire_me} uses these features in accordance with psychological indicators to tell the hirability of a candidate. Some multi-modal approaches like \cite{auto_proctor,intelligent} propose detailed pipelines with speech detection, gaze estimation, head-pose estimation, etc. features. However, this kind of extensive pipeline generally requires more computation power.
        
        Online proctoring systems supported by artificial intelligence have a remarkable effect on conducting exams in a reliable and secure environment \cite{auto_proctor}, \cite{unsupervised}. Our system includes a similar structure as the study \cite{auto_proctor} including another person detection and electronic device detection. Also, in \cite{auto_proctor} they used an eye tracker to understand where the candidate is looking during the exam. To catch more information from the face of a person dual vision camera is used in \cite{winarno}. However, it is not very feasible to provide an eye tracker or dual vision camera to all candidates in our system.


\section{Proposed Method}

    Our system aims to detect predefined cheating activities in a fully automated way just by processing on recorded video data of the test-taker or -as we call- \textit{candidate}. The system does not require any additional hardware except the webcam or any available camera. Moreover, there is no authentication phase where the candidate uploads some visual and textual information in advance. We only need the candidate to show themselves clearly in the first $\videostartprocessseconds{}$ seconds of the video to let us register the candidate's face for further process. All these features make our system quite fast and efficient as we delve into detail in the following sections.

    The remainder of this section describes the following topics:
    Dataset, Cheating detection pipeline, Video pre-processing, Candidate's face detection, Face recognition, Object detection, Face tracking, and Result analysis.

    \subsection{Dataset}
    
        We have designed and created our data collection and labeling strategy, because there is no publicly available dataset that fits to our system. We have used our online exam platform \footnote{https://www.talent-interview.com/} that measures the skills of participants for employers with hundreds of questions selected from a question pool. There are multiple-choice and free coding questions available in the pool. 
        
        We asked the participants to perform some intentional cheating actions. These actions include having another person in the exam environment other than the candidate, the candidate leaving the exam environment, using a mobile phone or laptop visibly, or any combination of these events throughout the video. Participants joined the experiments from different places like rooms, classrooms, or workplaces; used different camera angles and setups, been under different lighting conditions, etc. These variations made quite challenging cheating events to detect, for example, mobile phones were barely visible since participants hold them right next to their ears in a speaking position without showing the phone clearly to the camera. Or the candidate left the exam partially not showing their face just by showing a part of their body.

\begin{figure}[ht]
            \centering
            \includegraphics[scale=0.40]{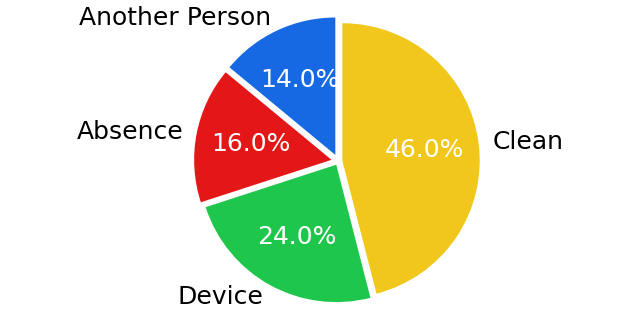}
            \caption{Cheating event distribution of our dataset with $\nvideos{}$ videos.}
            \label{fig:dataset_pie}
        \end{figure}

        From Huawei Turkey R\&D Center $\nparticipants{}$ participants with $\nvideos{}$ videos joined our experiments. As mentioned earlier, most of them 
        were asked to perform a cheating event deliberately, even though the total number of cheating events is not very realistic just to measure our performance with a lot more cheating events. Videos range from 10 minutes to 25 minutes.
        Two annotators watched and labeled cheating intervals in seconds (start second, end second). Each event was labeled separately. For instance, if \textit{absence} and \textit{another person} events happened concurrently, they are both annotated as two events. For \textit{another person} event, the face or body of the other person must be visible to be labeled, Similarly, for \textit{absence} event, the candidate's face and body together should be out of the scene. For \textit{device} event, using a mobile phone or laptop without showing the camera was not counted as a cheating event, however, even if a small part of the device is visible then it was annotated. A pie chart of our dataset that shows the counts of cheating events is below \ref{fig:dataset_pie}.
        
        As we said earlier, even slight differences in camera angles, lighting, visibility rates, etc. create a lot of variations in events. Other diversities emerged with the occurrence of the cheating events in different ways. For instance, sometimes a candidate leaves the exam environment and some other person becomes visible for a brief time. Or, a candidate might use an electronic device just by showing a part of its cover side. These problematic cases create controversies in both the labeling and testing phases.

    \subsection{Cheating detection pipeline}

        There are some assumptions about an ideal online interview system to define cheating events that we attempt to detect. First, we presume there must be a person whose face appears throughout the video and he or she is called as \textit{Candidate}. We define a legitimate online interview video where no cheating events are observed as follows:

        \begin{itemize}
            \item Only the candidate must appear throughout the video.
            \item Any use of a mobile phone and laptop is forbidden. However, we count this as cheating if it only appears prominently in the video.
            \item The candidate should not leave the environment for more than 5\% of the total length of the video.
        \end{itemize}

        Violating any of the above cases corresponds cheating activities as \textit{Another person}, \textit{Device}, and \textit{Absence} respectively, and thereby video is labelled as \textit{Suspicious}. On the contrary, if none of the three cheating activities is violated, then the overall evaluation of the video would be \textit{Clean}. We prefer the label \textit{Suspicious} instead of something like \textit{Cheating} due to the subjective nature of cheating events.
        
\begin{figure*}[ht]
            \centering
            \includegraphics[width=0.99\linewidth, height=4cm]{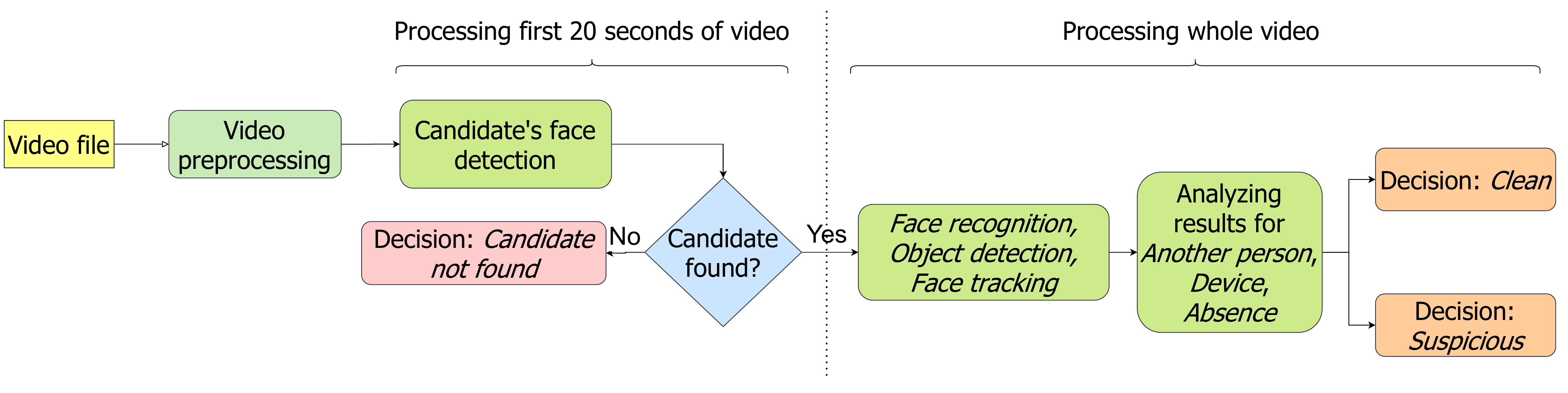}
            \caption{Proposed cheating detection pipeline. 1) Video file is preprocessed for the standard codec, constant fps, and resolution values. 2) In the first 20 seconds of the video, the candidate's face is detected and registered for further analysis. If the candidate's face cannot be found, then the video will not be processed further. 3) The whole video is processed traversing each frame with face detection, face tracking, and object detection modules. The result of each frame is reported in a table. 4) Table is analyzed to decide each cheating event field \textit{Another person}, \textit{Device} and \textit{Absence}. 5) Each field is labeled as either \textit{Clean} or \textit{Suspicious}. 6) If at least one field is labeled as \textit{Suspicious} then the final result is \textit{Suspicious} as well, otherwise, it is \textit{Clean}.}
            \label{fig:pipeline}
        \end{figure*}

        Our anti-cheating pipeline aims to decide on each of these three events and it consists of five main parts: \textit{Video pre-processing}, \textit{Candidate's face detection}, \textit{Face recognition}, \textit{Object detection} and \textit{Result analysis}. Details of the pipeline can be seen in Figure \ref{fig:pipeline}.
        
    \subsection{Video pre-processing}

        Video pre-processing is critical because it helps subsequent steps to run faster. Besides, it adjusts all videos in a standard decoding format with fixed resolution and fps values.
        
        We set our videos at a constant $3$ fps rate. It may seem a bit lower for a standard frame sequence, but it is still adequate to extract meaningful cheating relations between frames. Also, we used $400$ pixel width for frames while keeping original aspect ratios of them.

    \subsection{Candidate's face detection}

        After obtaining pre-processed video, the very first step is to locate and register the candidate's face encodings for further evaluations.
        
        We have employed two main algorithms for face location: Histogram of Oriented Gradients (HOG) \cite{hog} based Support Vector Machine (SVM) detector and Convolutional Neural Network (CNN) based detector. We have utilized pre-trained models provided by dlib \cite{dlib} library for both.
        
        For further cheating analysis of the video, we should decide on the candidate's face and register its face encodings. At this stage, we have utilized only HOG detector. In the first $20$ seconds of the video, we are locating all the faces in each frame. But an auxiliary head-pose estimation algorithm checks the face angle towards the screen. If the angles between the x and y-axis of the screen are out of pre-defined thresholds, then those face images are not accepted as candidate's face images. Then, we are saving encoded parameters of all detected faces. These parameters are $128D$ vectors acquired from the HOG-based SVM classifier model registered for further face comparisons in the video.

    \subsection{Face recognition}

        During the process of the whole video, we process each frame to detect the presence of the candidate's face, to count the faces and bodies, and to detect the appearance of an electronic device.
        
        While traversing the frames, for each located face we make a comparison to the encodings of the registered face to the candidate's face. As mentioned before, registered frames were collected from the first 20 seconds of the video and we call them $F_{s}$ where $s$ indicates the frame index in the first 20 seconds. Traversed frames are called $F_{t}$ where $t$ indicates the frame indices of the whole video. For any frame, $F_{t}$, face distance is computed as below,
        
        \begin{equation}\label{frame_face_distance_formula}
            min({||D_{1}||}_F, {||D_{2}||}_F, ..., {||D_{s}||}_F, ..., {||D_{n}||}_F)
        \end{equation}
        
        where ${||D_{s}||}_F$ is the Frobenius norm of element-wise difference of $F_{s}$ and $F_{t}$ as below
        
        \begin{equation}\label{single_frame_distance_formula}
            {||D_{s}||}_F = {||F_{s} - F_{t}||}_F
        \end{equation}
        
        which can also be represented as
        
        \begin{equation}\label{frobenius_norm}
            {||D_{s}||}_F = \sqrt{\sum_{i=1}^{128}|{F_s}_{i} - {F_t}_{i}|^2}
        \end{equation}

        As in the Equation \ref{frame_face_distance_formula} the distance to the closest face among all registered faces is accepted as the face distance. Here, we employed another simple but effective strategy for partially seen faces, which we call \textit{partial face matching}. In Equation \ref{single_frame_distance_formula}, when we compute the distance between $F_s$ and $F_t$ we make an assumption that the values lower than $\epsilon=10^{-2}$ in $F_t$ are coming from unseen or imperceptible features of the detected face. Thereby we discard the distances that come from these small values, by equalizing $F_t$ with $F_s$ at the same indices in $128D$ face encodings vector. In other words, small values in $F_t$ are not taken into account in face distance calculation. This assumption prevents many false alarm cases by giving lower face distances when a candidate is present but their face is only partially seen.
        
        Outlier face distances are determined with a constant thresholding mechanism. Those frames with face distances that exceed the threshold which is set as $0.65$ in our experiments are nominated as the frames with another person. Afterward, we link those frames to each other depending on their closeness in the sequence to decide how many times another person appeared in the video. If those frames connect (if they are close to each other), then each sequence is counted as $1$ appearance of another person.

    \subsection{Object detection}

        For the purpose of detecting person bodies and devices, we have employed an object detection Mobilenet-SSD model \cite{mobilenet-ssd} trained with 3 classes (body, mobile phone, and laptop) of COCO2017 dataset \cite{coco} and run on each traversed frame $F_{t}$ as well.
        
        Similar to face detection, object detection is determined through a thresholding mechanism. But unlike face detection, since we obtain a confidence value from the Mobilenet-SSD model, we threshold these values to decide cheating activities instead of distance values. For body and device detection, the thresholds are $0.65$ and $0.30$ respectively in the experiments. These are optimized threshold values according to our experiments to have a balance between true detections and false alarms.
        
        For body detection, we collect body counts and confidence values from each frame. The frames with a body count $1$ but no candidate's face found or with body count is more than 1 even though the candidate's face is found are the nominated ones. Similar to the face detection part, those nominated frames are linked to each other depending on their distances in the sequence to decide how many times multiple bodies appeared in the video. If it is at least $1$ time, then we decide another person appeared.
        
        Our adapted Mobilenet-SSD with 3 classes gives out results for each class (body, mobile phone, laptop). Therefore as in the body detection, confidences are thresholded. Those frames that exceed the threshold are linked to each other to decide how many times a device appeared. If it is at least $1$ time, then we decide that the device appeared.

    \subsection{Face tracking}

        For better face comparison we also have employed a face tracking algorithm which is the correlation tracker \cite{correlation_tracker}. Similar to face detection algorithms, the implementation of it is provided by dlib library \cite{dlib}.
        
        Face tracking assists face detection at frames where the detection algorithm fails. These cases often occur where the candidate temporarily blocks his or her face or the face is visible with an odd angle. In this case, we prioritize the result of face tracking. However, in extreme cases where the candidate leaves the environment or the face is blocked so that the face tracking confidence drops prominently, face tracking is deactivated.
        
        The other critical case occurs when the location of the faces detected by face detection and face tracking algorithms do not overlap. If they are apart from each other too much, relative to the length of the frame diagonal, then we mark the frame with more than one face to be evaluated as a frame with another person.

    \subsection{Result analysis}

        We record the results of each frame in a table to be analyzed after the processing completes. The table includes results of each component of anti-cheating features which are face detection, body detection, and device detection.
        
        Face and body detection results are merged into \textit{Another person} decision. If the count of one of them is at least $1$ then the \textit{Another person} field is reported as \textit{Suspicious}, otherwise it is \textit{Clean}. As we said earlier, \textit{Device} field is reported from the direct result of device detection as either \textit{Clean} or \textit{Suspicious}.
        
        To decide if \textit{Absence} occurred or not, we have a $5\%$ upper limit. If the candidate`s face is not found in a frame with body count being $0$ and the ratio of such frames exceeds $5\%$ limit, we decide to report \textit{Absence} is \textit{Suspicious}. Otherwise, it is \textit{Clean}. 
        
        As explained before, if at least one of these three events is reported as \textit{Suspicious}, then the overall decision for the video is \textit{Suspicious} as well, otherwise, it is \textit{Clean}.

        \begin{figure*}[h]
            \centering
            \includegraphics[scale=0.1]{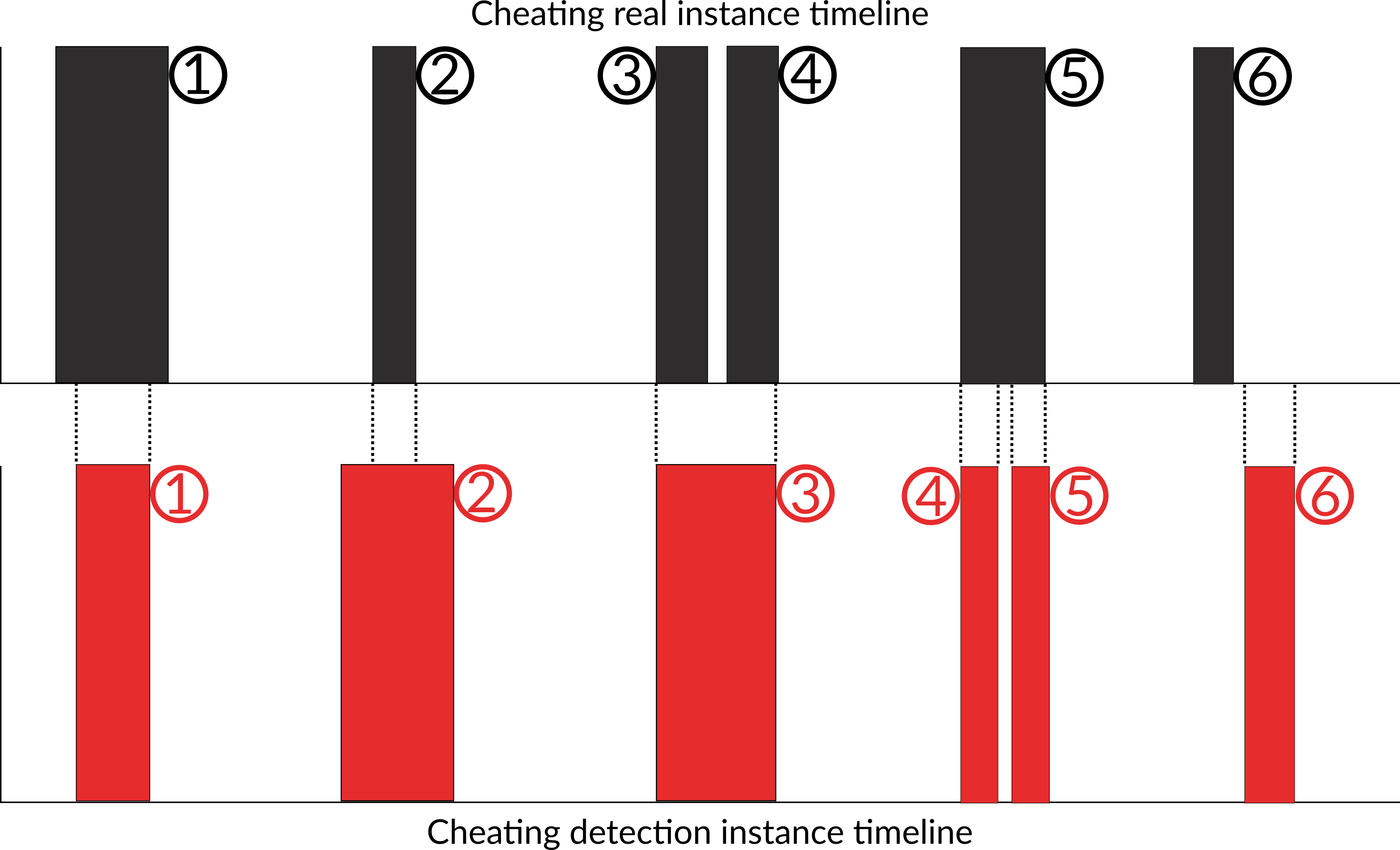}
            \caption{Sample cheating and detection instances on separate timelines. Instances $1$ and $2$ in both timeline show the matching cases even though the detection instances do not quite overlap with real instances. Cheating instance $6$ is a missed and detection instance $6$ is a false alarm case. Cheating instances $3$ and $4$ combined, and instance $5$ individually shows the case where a careful matching with predictions is required. In these cases, we match the cheating instance with the closest detection instance and check their IOU values for calculating true detections. Similarly, to be able to decide a false alarm case, we match the detection instance with the closest cheating instance. Therefore, same detection instance may be counted as true detection more than once, or there may be multiple false alarms for the same cheating instance. All these cases emphasize that instance based metric is a bit difficult to measure the real performance of our system since it emerges a matching complexity.}
            \label{fig:plot_instance}
        \end{figure*}

\section{Experimental Results}

    We have conducted experiments on a private dataset that we have acquired and labeled from online interviews of $\nparticipants{}$ different participants which in total has $\nvideos{}$ different interview videos with lengths ranging from $10$ minutes to $25$ minutes. Most of the videos contain intentional cheating activities that we have defined earlier such as another person's appearance, electronic device usage, or absence of the candidate.
   
    Each video was watched and labeled according to their cheating events. If at least one of three events (\textit{Another person}, \textit{Device}, \textit{Absence}) appears to be prominent, then the video is labelled as \textit{Suspicious} with the associated cheating event to be \textit{Suspicious} as well.

    \subsection{Metrics}
    
        As in \cite{auto_proctor}, we have utilized instance and segment-based metrics to measure our performance. Additionally, we have measured overall accuracies by video-based labels. We calculate accuracies by two ratios: True Detection Rate (TDR) and False Alarm Rate (FAR). TDR is the ratio of the number of correct predictions to the number of all positive labels which measures how many of the cheating actions are predicted correctly. On the other side, FAR is the ratio of the number of false predictions to the number of all positive predictions which measures how many predictions are wrong.
        
        \subsubsection{Instance-based}
        
            A cheating instance is a rectangle region with black borders and the detection instance is the red region as drawn in Figure \ref{fig:plot_instance}. For matching regions, we have used IOU (Intersection over Union) ratio. So, for a cheating instance if the detection instance with the highest IOU of all detection instances is above $\IOU$, then it counts as a true detection. Similarly, for a detection instance if the cheating instance with the highest IOU of all cheating instances is below \IOU, then it counts as a false alarm. In Figure \ref{fig:plot_instance}, TDR is $1.0$ whereas FAR $0.2$ by instance metric.
            
            Table \ref{table:instance_metrics} shows the performance of our system by instance-based metric on our private dataset.
            
            \begin{table}[h]
                \centering
                \caption{Instance TDR and FAR values on our dataset.}
                \label{table:instance_metrics}
                \begin{tabular}{l|l|l}
                    & Instance TDR & Instance FAR \\ \hline
                    Another person & 0.405 & 0.047 \\
                    Device & 0.567 & 0.017 \\
                    Absence & 0.587 & 0.205 \\
                \end{tabular}
            \end{table}
        
        
        \subsubsection{Segment-based}

            Frame sequences are segmented by constant segment lengths by their cheating labels and detection results. We have two frame sequences as cheating sequence and detection sequence. Their elements are binary and they represent cheating labels and predictions respectively. 
            
            For each segment, these two array elements are compared. Array elements correspond to frames. A segment is a cheating segment when the ratio of positive elements to the length of the segment array is above $\segmentmatchrate$ in the cheating sequence. Similarly, in the detection sequence, such segments are detection segments.
            
            When a segment is a cheating segment, if the corresponding segment is a detection segment in the detection sequence, then we have a true detection. Oppositely, when a segment is a detection segment if the corresponding segment is not a cheating segment in a cheating sequence, then we have a false alarm.
            
            Table \ref{table:segment1_metrics} and \ref{table:segment3_metrics} show the performance of our system by segment based metric on our private dataset. Segment lengths are $1$ and $3$ seconds or $3$ or $9$ frames since experiments were at 3 FPS.

            \begin{table}[ht]
                \centering
                \caption{Segment (1 sec) TDR and FAR values on our dataset.}
                \label{table:segment1_metrics}
                \begin{tabular}{l|l|l}
                    & Segment (1 sec) TDR & Segment (1 sec) FAR \\ \hline
                    Another person & 0.316 & 0.023 \\
                    Device & 0.389 & 0.024 \\
                    Absence & 0.625 & 0.189 \\
                \end{tabular}
            \end{table}
        
            \begin{table}[ht]
                \centering
                \caption{Segment (3 sec) TDR and FAR values on our dataset.}
                \label{table:segment3_metrics}
                \begin{tabular}{l|l|l}
                    & Segment (3 sec) TDR & Segment (3 sec) FAR \\ \hline
                    Another person & 0.317 & 0.023 \\
                    Device & 0.410 & 0.023 \\
                    Absence & 0.741 & 0.184 \\
                \end{tabular}
            \end{table}

            

        \begin{figure*}[h]
            \centering
            \includegraphics[width=0.99\linewidth, height=7.2cm]{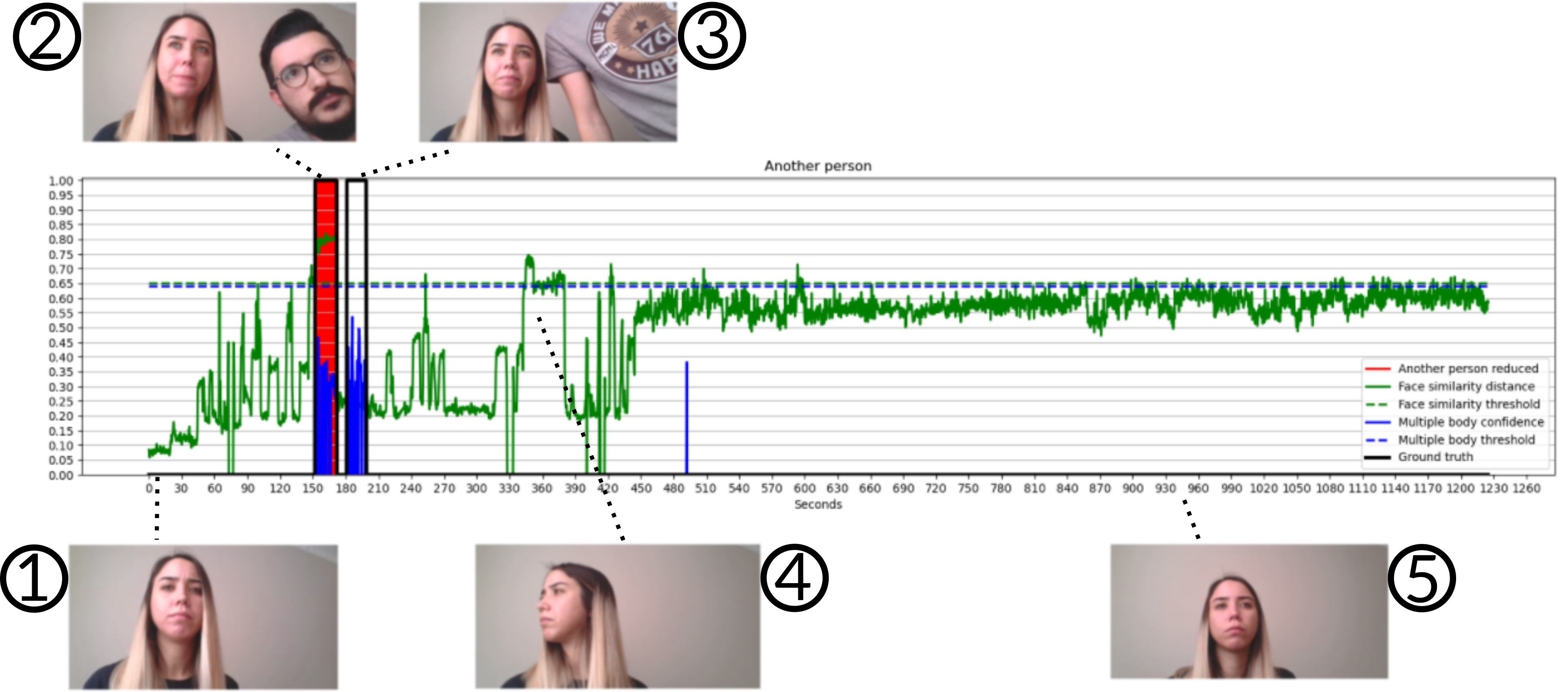}
            \caption{Sequence analysis of \textit{Another person} instances of a sample video with some corresponding frames on top and bottom is above. Black and red colors specify the ground truth and detection instances respectively. Green and blue lines indicate face distance and multiple body confidence values. Dashed lines at 0.65 indicate their threshold values. Frame $1$ is the registered candidate face. In frame $2$ there is another person and it is detected, whereas in $3$ other person's face is not visible and multiple body confidence is below the threshold, and it is a miss. Frame $4$ shows a case where the candidate's face angle causes the face distance to increase. However, the face tracking model prevents a false alarm. In the second half of the video, face distance goes quite large due to the different face position of the candidate as in frame $5$ than frame $1$. Nevertheless, there is no false alarm at this part.}
            \label{fig:plot_ap}
        \end{figure*}
        
        \begin{figure*}[h]
            \centering
            \includegraphics[width=0.99\linewidth, height=7.2cm]{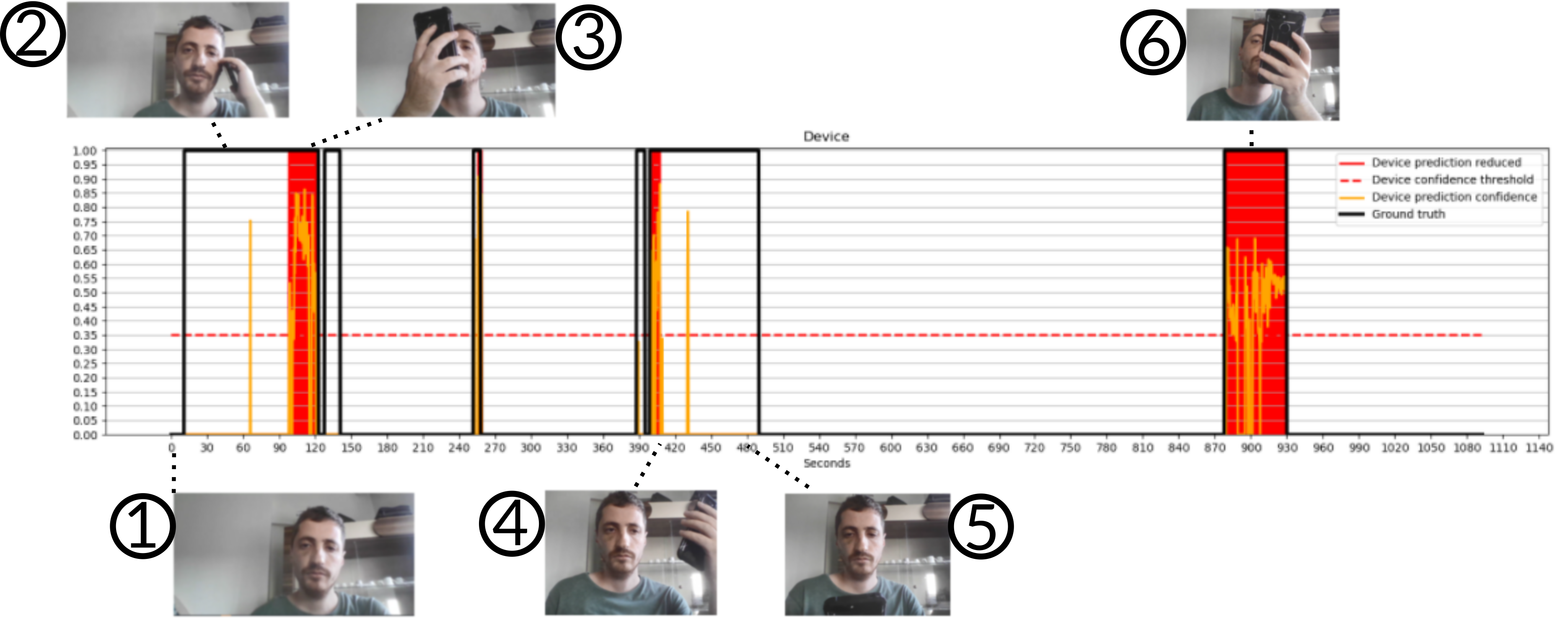}
            \caption{Sequence analysis of \textit{Device} instances of a sample video with some corresponding frames on top and bottom is above. Black and red colors specify the ground truth and detection instances respectively. Orange lines are raw confidences of our object detection model and are converted into red regions through sequence analysis. Dashed lines indicate the threshold value. Frame $1$ is the registered candidate face. In frame $2$, even though phone call action is obvious, the phone itself is barely visible due to bad angles. In the ending of the same cheating instance, the candidate holds the phone at a better angle as in frame $3$, and there we detect it. Frames $4$ and $5$ show a very similar case as $2$ and $3$. In frame $6$, the phone is apparent once again and the detection is satisfying.}
            \label{fig:plot_device}
        \end{figure*}
        
        \begin{figure*}[h]
            \centering
            \includegraphics[width=0.99\linewidth, height=6cm]{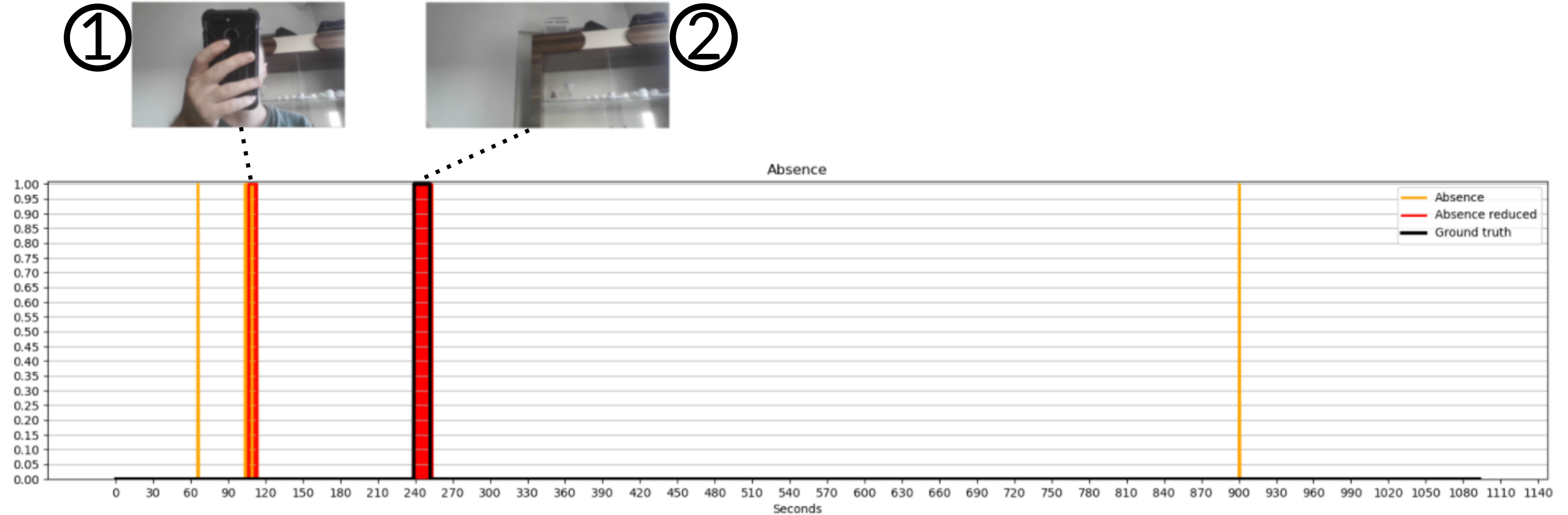}
            \caption{Sequence analysis of \textit{Absence} instances of the same video as in Figure \ref{fig:plot_device} with some corresponding frames on top is above. Black and red colors specify the ground truth and detection instances respectively. Orange regions are raw prediction results and eliminated through the sequence analysis part. In frame $1$, the device is blocking the face and a false absence alarm occurs. On the other hand, in frame $2$, the real absence case is detected correctly.}
            \label{fig:plot_absence}
        \end{figure*}

        \subsubsection{Video-based}
        
            Each video sample has one binary label by their cheating events (\textit{Another person}, \textit{Absence}, and \textit{Device}). If there is at least one cheating instance and at least one detection instance throughout the frame sequence, this is a true detection. Similarly, if there is no cheating instance and we have at least one cheating detection, it is a false alarm.
            
            Below in Table \ref{table:overall_prf_metrics}, precision-recall-F1 metrics are presented. Positive labels are presumed as \textit{Suspicious} whereas negative labels are presumed as \textit{Clean}. Therefore, true-positive and true-negative cases are correctly predicted \textit{Suspicious} and \textit{Clean} ones respectively, whereas false-positive and false-negative cases are the ones incorrectly predicted as \textit{Suspicious} and \textit{Clean} cases respectively.
        
            \begin{table}[ht]
                \centering
                \caption{Overall performance by precision, recall and F1 metrics with on our dataset.}
                \label{table:overall_prf_metrics}
                \begin{tabular}{l|l|l|l}
                    & Precision & Recall & F1 \\ \hline
                    Another person & 0.55 & 0.86 & 0.67 \\
                    Device & 1.0 & 0.83 & 0.91 \\
                    Absence & 0.89 & 0.89 & 0.89 \\
                    \textbf{Overall cheating} & \textbf{0.90} & \textbf{0.86} & \textbf{0.88}
                \end{tabular}
            \end{table}

    Among all three metrics presented above, the instance metric is the most unclear one since it requires the matching of cheating and detection instances depending on their intersections. Besides, it is very controversial and parameter-dependent that how much intersection or union is sufficient for the matching of instances. Thereby, even though we have presented an instance-based metric, we do not consider it as an appropriate metric for this kind of application.
    
    The segment-based metric is a more clear and lucid metric compared to instance based one. It does not require a matching decision and does not take intersection areas into account. However, it is still very changeable depending on segment length choice.
    
    Although using video-based metric decreases the number of cheating events in the dataset since detecting individual instances are irrelevant, we believe it is a more suitable predilection for measuring system accuracy. Because our system is designed and optimized for detecting cheating activities as fast as possible without exhaustive processes. This makes measuring the performance with a video-based metric more important for our system.
    
    In Figures \ref{fig:plot_ap}, \ref{fig:plot_device} and \ref{fig:plot_absence} there are sequence plots from two different sample videos where horizontal axes are the seconds, and vertical axes define a range between $[0, 1]$ for ground truths and predictions as well as raw predictions of our system. Raw predictions are indicated with orange and they are converted into red regions that are our final prediction instances. There are both success and fail cases shown in Figures \ref{fig:plot_ap}, \ref{fig:plot_device}, and \ref{fig:plot_absence}. For example, if a candidate's face is shown with an odd angle, or another person's face or body is not clearly seen, then our system might produce wrong predictions. 

    
    Unfortunately, our private dataset prevents us from making a direct comparison to other similar works' performances since our labeling and performance evaluation approaches do not quite overlap with these works \cite{auto_proctor,intelligent}. Nevertheless, individual field and overall cheating performances show promising results. 
    
    Relatively higher false-positive rates due to unrecognized candidate faces cause lower precision scores in \textit{Another person} field. Also in the \textit{Device} field, there are minor false-negative cases since candidates are mostly in an effort to hide their mobile phone usage so that the device is not visible prominently.
    
    \subsection{Hardware performance}

        We have used Python \footnote{python.org/downloads} language for the implementation of the whole system. All tests were conducted on a workstation with Intel i9 20 CPU cores at 3.60GHz with 64 GB RAM. Operating system was Ubuntu 18.04 \footnote{releases.ubuntu.com/18.04}. 
        Besides, in order to have an easily deployable system to a wide range of machines, we have packaged our environment with Pyinstaller \footnote{pyinstaller.org} tool, so that all of the required libraries and face/object detection models could be bundled together. The bundle size is around $900$ MB. Processing times of all components can be seen in Table \ref{table:process}.
        
        \begin{table}[ht]
            \centering
            \caption{Processing times of all components of our system.}
            \label{table:process}
            \begin{tabular}{l|l}
                & FPS \\ \hline
                Video pre-processing & \multicolumn{1}{r}{$16$} \\
                Candidate's face detection & \multicolumn{1}{r}{$498$} \\
                Face recognition & \multicolumn{1}{r}{$14$} \\
                Object detection & \multicolumn{1}{r}{$47$} \\
                Result analysis & \multicolumn{1}{r}{$1.5x10^{6}$} \\
                \textbf{Overall} & \multicolumn{1}{r}{$\textbf{4.9}$}
            \end{tabular}
        \end{table}

        Considering our system can work at constant 3 FPS videos, the overall speed of 4.9 FPS in Table \ref{table:process} means that the system surpasses real-time speed. In other words, a 1 minute long video is being processed around $~37$ seconds.

\section{Discussion}

    In this work, we have focused on designing an efficient pipeline for online interviews and exams with state-of-the-art computer vision and video processing techniques. Since having a fast pipeline is proposed, we had to limit the system inputs and outputs. Inputs are constrained with visual data and audio data is ignored. And the system outputs of the cheating analysis are confined just by three cheating events that are \textit{Another person}, \textit{Device} and \textit{Absence}. This makes our system more like a computer vision application.
    
    Even though we are only processing visual data, we have employed a combination of classical image and video processing techniques with state-of-the-art deep learning models in an easily deployable and fast algorithmic pipeline. We also claim that the components of the pipeline can be detached and used separately since most of them do not depend on each other. For example, if \textit{Absence} case is not important for an application, that part would be discarded from \textit{Result analysis} phase. Or if one is only interested in the usage of electronic devices in an exam environment, only the object detection module can run, and be analyzed accordingly.
    
    Another advantage of our system is that it does not require any additional hardware or any data other than a video file. All authentication and processing steps rely on one single video file that is recorded during the exam. On the other hand, an obvious weakness of such an approach is that the inability to detect any cheating action that is out of the scene. Without using auxiliary hardware such as a wearcam as in \cite{auto_proctor}, it is not possible to have all the information of the exam environment. Nonetheless, with this limited environment information and relatively low processing value (3 FPS), our system promises a reliable assurance at detecting main cheating activities. Moreover, not using those auxiliary hardware makes the system simple, easily applicable, and fast. 
    
    Besides, this system can be considered as an application that works collaboratively with human proctors. Since detected cheating event seconds are known, those parts of the video can be trimmed and be checked by the proctor. Such usage is more desirable compared to checking the whole video for a possible cheating event.
    
    As mentioned earlier, Figures \ref{fig:plot_ap}, \ref{fig:plot_device} and \ref{fig:plot_absence} show some success and fail cases together. Since we have access only to the front side view of both candidate and exam environment, if the interested cheating event is not happening at good visual angles, our system might fail at detecting those cases or produce false alarms.
    
    Further works that can improve the performance of our system would be employing a voice analyzing module as most of the cheating events happen by only speaking or other cheating events are accompanied by suspicious voice activity. The other improvement is that adding a gaze tracking module similar to \cite{customers,gaze_estimation} to analyze the eye movements of the candidates. We believe that these two modules can greatly contribute to the accuracy of our system.

\section{Conclusion}

    This paper presents a cheating detection pipeline. The main goal of the system is to provide a reliable and secure exam environment for online interviews and exams. As input, it only requires a small-sized candidate's video that can be recorded by an integrated webcam. Therefore it is an easily applicable and fast system. The pipeline includes face detection, face recognition, face tracking, and object detection components. We determined three main cheating activities that are Another person, Device, and Absence. We conducted the experiments on a private dataset that consists of $\nvideos{}$ videos with real-life cheating activities. As a measurement, we utilized three different metrics that are instance-based, segment-based, and video-based metrics. As a result, we obtained $0.88$ F1 score by video-based metric. As a further step, we might expand our work with voice analysis and gaze estimation features.


%

%



\ifCLASSOPTIONcaptionsoff
  \newpage
\fi





\end{document}